# Improving Early Prediction of Type 2 Diabetes Mellitus with ECG-DiaNet: A Multimodal Neural Network Leveraging Electrocardiograms and Clinical Risk Factors

Farida Mohsen[1,*] and Zubair Shah[1,*]

[1]College of Science and Engineering, Hamad Bin Khalifa University, Doha, Qatar

## ABSTRACT

Type 2 Diabetes Mellitus (T2DM) remains a significant global health challenge, underscoring the need for early and accurate risk prediction tools to enable timely interventions. This study introduces ECG-DiaNet, a multimodal deep learning model that integrates electrocardiogram (ECG) features with established clinical risk factors (CRFs) to improve the prediction of T2DM onset. Using data from the Qatar Biobank (QBB), we compared ECG-DiaNet against unimodal models based solely on ECG or CRFs. A development cohort ($n = 2043$) was utilized for model training and internal validation, while a separate longitudinal cohort ($n = 395$) with a five-year follow-up served as the test set. ECG-DiaNet demonstrated superior predictive performance, achieving a higher area under the receiver operating characteristic curve (AUROC) compared to the CRF-only model ($0.845$ vs. $0.8217$), which was statistically significant based on the DeLong test ($p < 0.001$), thus highlighting the added predictive value of incorporating ECG signals. Reclassification metrics reinforced these improvements, with a significant Net Reclassification Improvement (NRI = $0.0153$, $p < 0.001$) and Integrated Discrimination Improvement (IDI = $0.0482$, $p = 0.0099$), confirming the enhanced risk stratification. Furthermore, stratifying participants into Low-, Medium-, and High-risk categories revealed that ECG-DiaNet provided a superior positive predictive value (PPV) in the high-risk group compared to CRF-only models. The non-invasive nature and widespread accessibility of ECG technology support the clinical feasibility of ECG-DiaNet, particularly in diverse healthcare settings, including primary care and community-based health centers. By leveraging both cardiac electrophysiological insights and systemic risk profiles, ECG-DiaNet addresses the multifactorial nature of T2DM and shows promise for advancing precision healthcare strategies. These findings underscore the potential of multimodal deep learning models to improve early detection and prevention efforts for T2DM, especially within underrepresented Middle Eastern populations.

## Introduction

The global prevalence of Type 2 Diabetes Mellitus (T2DM) was estimated at 9.8% in 2021[1], significantly lower than the 18.1% reported in countries within the Middle East and North Africa region. . In Qatar, T2DM prevalence is estimated at 14% to 17%[2], aligning with trends observed in other Middle Eastern populations. This elevated prevalence compared to Western populations increases the risk of T2DM-related complications, including cardiovascular disease (CVD), retinopathy, nephropathy, and elevated mortality. Early identification of high-risk individuals is therefore of critical clinical importance, as targeted lifestyle and behavioral interventions can significantly reduce disease onset. Preventive measures have demonstrated reductions in progression to T2DM by up to 58%, emphasizing the importance of risk prediction in precision medicine[3,4]. While established tools, such as the Framingham Risk Score and Finnish Diabetes Risk Score, rely on traditional factors such as age, body mass index (BMI), and blood pressure[5,6], researchers have continuously sought to improve these models by integrating new and more informative risk factors.

CVD is one of the most common complications of T2DM, as over one-third of patients with diabetes also go on to develop CVD[7]. The electrocardiogram (ECG), a widely available and non-invasive diagnostic tool, captures valuable information on cardiac electrical activity closely linked to T2DM-related cardiovascular changes. Given the established interplay between cardiovascular and metabolic health, ECG data is promising for identifying diabetes-related cardiac abnormalities and early indicators of disease progression[8,9]. Furthermore, given the close association between T2DM and CVD, some studies have utilized machine learning models to analyze ECG data for detecting existing T2DM. For example, Lin et al.[10] developed an ECG-based model to predict HbA1c levels using a one-dimensional residual module. Similarly, other research has demonstrated the effectiveness of ECG-based approaches for diabetes detection[8,9].

While significant progress has been made, most existing research has concentrated on identifying current cases of T2DM, with limited investigation into using ECG data to predict future T2DM risk. A notable exception is the study by Kim et al., which introduced the concept of a "diabetic ECG" through pioneering work in South Korea. This study utilized deep learning to analyze 12-lead ECG data independently of traditional glycemic markers, identifying patterns associated with an increased risk of developing diabetes. However, the focus on a South Korean population raises questions about the applicability of these findings to other groups, particularly those with different genetic, environmental, and lifestyle backgrounds, such as Middle Eastern populations. Additionally, the study incorporated demographic and certain morbidity features alongside ECG data. To date, the potential benefits of integrating ECG features with established clinical risk factors in multimodal predictive models have not been thoroughly explored, especially in Middle Eastern populations like that of Qatar. Combining ECG data with clinical risk factors could leverage their complementary strengths to enhance the accuracy of T2DM risk predictions.

This study examines the utility of 12-lead ECG data for T2DM risk prediction over a median five-year follow-up period, using data from the Qatar Biobank (QBB)[11], thereby addressing a gap in long-term T2DM risk prediction research. We introduce ECG-DiaNet, a multimodal neural network-based model that integrates ECG-derived features with established CRFs, leveraging their complementary information to enhance predictive accuracy. We rigorously compare this multimodal approach against unimodal models based solely on ECG or CRFs. By centering our analysis on a Middle Eastern cohort, specifically the Qatari population, we underscore the importance of population diversity in predictive modeling. To the best of our knowledge, this is the first study to employ a multimodal model combining ECG features and CRFs for T2DM risk prediction, especially in such an underrepresented population. This non-invasive, scalable approach holds significant promise for improving early detection and risk stratification efforts, thereby contributing new insights through the integration of longitudinal data from an underrepresented population.

## Results

In this study, we developed and evaluated ECG-DiaNet, a multimodal neural network integrating ECG features with CRFs for T2DM risk prediction. Two cohorts were derived from the QBB: a development cohort for model training and validation, and a longitudinal test cohort for assessing future T2DM risk prediction. The development cohort consisted of 2,043 participants, with a mean age of 46.59±14.80 years, and a predominance of male participants (68.3%). The longitudinal test cohort included 395 non-diabetic individuals, slightly younger with a mean age of 37.59±11.24 years, and a more balanced gender distribution (52.6% male). Figure 1 illustrates the study design, and Table 1 provides detailed baseline characteristics of both cohorts. Three model configurations were evaluated: a CRF-only baseline model, ECG-only model, and the multimodal ECG-DiaNet model combining both ECG features and CRFs. To ensure robustness, five-fold cross-validation was performed with 1,000 bootstraps for each test fold during model development (see Methods). The final models were then trained on the entire development cohort and evaluated on the longitudinal test cohort to assess their predictive accuracy for T2DM risk.

### Cross-Validation Performance

Table 2 provides the average performance metrics across cross-validation folds for the unimodal ECG-based and CRF-based models, as well as the multimodal ECG-DiaNet model. The CRF-only baseline model demonstrated excellent performance, achieving an AUROC of 0.9841 (95% CI: 0.9839–0.9843), with high sensitivity (0.9484) and specificity (0.9612). These results underscore the robustness of CRFs for T2DM detection and establish the CRF-only model as a strong and reliable diagnostic baseline, highlighting the utility of clinical risk factors in this context. In contrast, the ECG-only model demonstrated moderate capability, achieving an AUROC of 0.8433 (95% CI: 0.8425–0.8440), indicating that ECG-derived features capture relevant cardiac electrical patterns associated with T2DM. While the model showed reasonable sensitivity (0.8059), its specificity (0.6951) was lower, suggesting a higher false-positive rate compared to the CRF-only model. The F1-score (0.7812) and AUPRC (0.8665) further emphasize its limitations as a standalone diagnostic tool. Despite its moderate performance, the ECG-only model highlights the potential of ECG features as non-invasive indicators for T2DM detection, particularly in preliminary screening scenarios where traditional CRFs are unavailable. When ECG features were integrated with CRFs in the ECG-DiaNet model, performance improvements were observed across multiple metrics. ECG-DiaNet achieved an AUROC of 0.9852 (95% CI: 0.9851–0.9854), surpassing the standalone models. Sensitivity and specificity also improved slightly, reaching 0.9514 and 0.9645, respectively, compared to 0.9484 and 0.9612 for the CRF-only model. The F1-score of 0.9602 further reflected the model's improved diagnostic performance. These results demonstrate the synergistic value of integrating ECG features with CRFs, solidifying ECG-DiaNet as the most effective model for T2DM detection in this cross-validation analysis.

### T2DM Risk Prediction Evaluation

While the detection results demonstrated the models' ability to identify existing T2DM cases, the true utility of predictive modeling lies in its capacity to forecast long-term risks. Predicting T2DM risk allows for timely interventions, potentially mitigating disease progression and improving clinical outcomes. To assess this capability, the models were evaluated on the

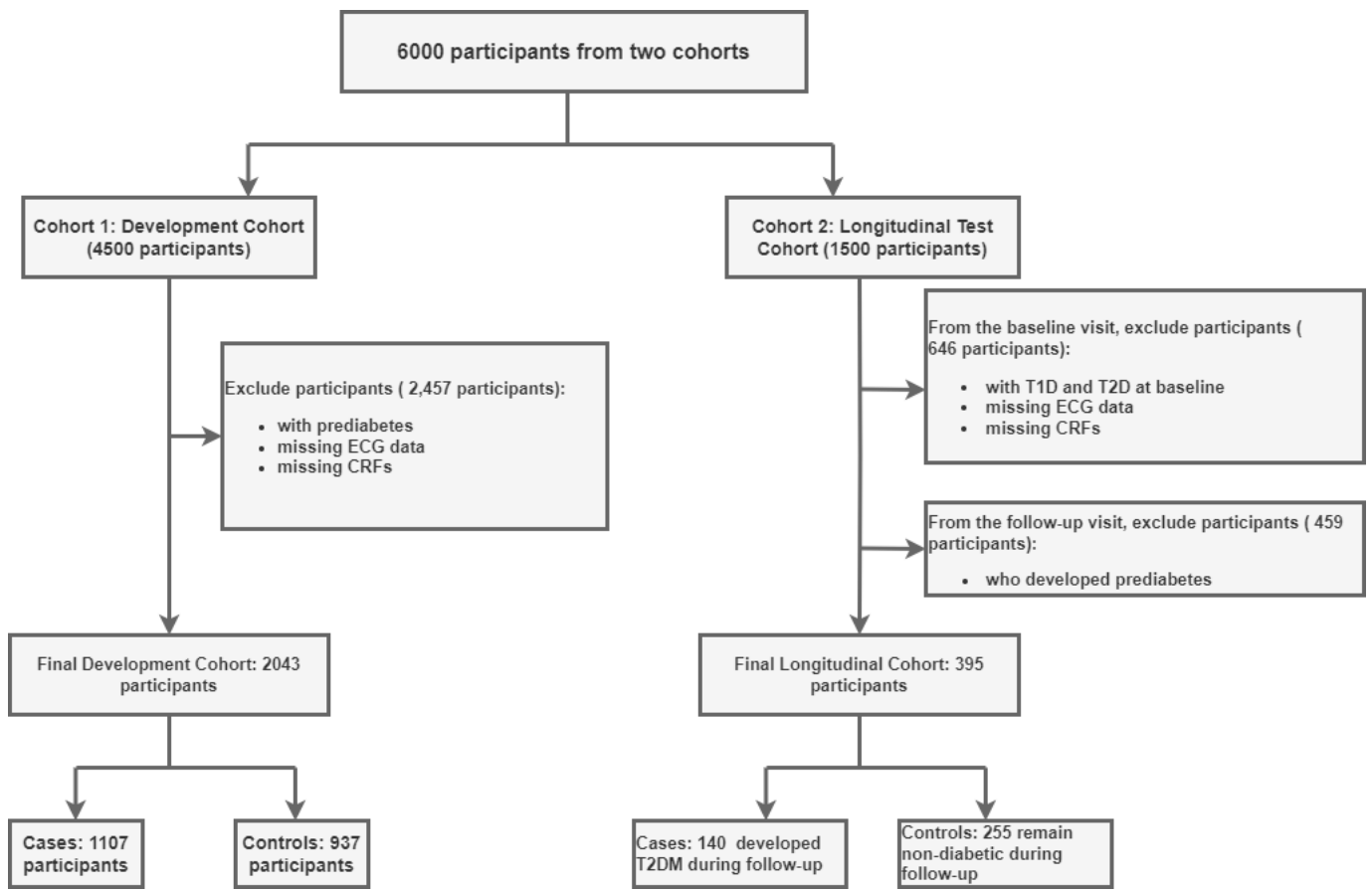

**Figure 1.** Study flowchart

**Table 1.** Characteristics of the Development and Test Cohorts

| Variable | Development Cohort (n = 2043) | Longitudinal Test Cohort (n = 395) |
|---|---|---|
| **Mean Age, years (SD)** | 46.59 (14.80) | 37.59 (11.45) |
| **Systolic BP, mmHg (SD)** | 118.37 (18.92) | 112.22 (13.40) |
| **Diastolic BP, mmHg (SD)** | 65.64 (10.00) | 66.32 (10.62) |
| **Waist Size, cm (SD)** | 88.22 (16.04) | 88.67 (14.79) |
| **BMI, kg/m² (SD)** | 27.86 (6.71) | 28.84 (5.83) |
| **ECG - QRS Duration (SD)** | 94.17 (12.49) | 94.51 (10.57) |
| **ECG - QT Interval (SD)** | 389.30 (27.22) | 390.59 (26.13) |
| **ECG - QT Corrected (SD)** | 403.14 (20.25) | 400.35 (18.46) |
| **ECG - PR Interval, (SD)** | 163.29 (23.09) | 159.73 (21.00) |
| **ECG - Average RR Interval (SD)** | 899.17 (138.42) | 927.10 (130.37) |
| **ECG - T Axis, degrees (SD)** | 38.01 (25.98) | 32.04 (20.88) |
| **Gender (Male)** | 1396 (68.3%) | 208 (52.6%) |
| **Number of Visits** | 1 | 2 |

**Note:** Continuous variables (e.g., Age, Systolic Blood Pressure (Systolic BP), Diastolic BP, BMI, Waist Size, ECG features) are presented as mean ± standard deviation. Categorical variables (Gender) are presented as n (%).

longitudinal test cohort, providing insights into their performance in identifying non-diabetic individuals at risk of developing T2DM during follow-up. The results, summarized in Table 3, highlight the predictive capabilities of the proposed models. The CRF-only model served as a baseline for evaluating the added value of integrating the ECG features into the predictive model. As shown in Table 3, this model achieved an AUROC of 0.8217 (95% CI: 0.8205–0.8229) and an AUPRC of 0.699, demonstrating moderate discriminatory power. It achieved a high sensitivity (0.764) and specificity (0.744), reflecting a strong ability to correctly identify T2DM cases while minimizing false positives. In addition, the F1-score (0.684) underscores the model's balanced performance. In comparison, the ECG-only model exhibited a moderate predictive capability, with an AUROC of 0.6747 (95% CI: 0.6730–0.6764). The sensitivity (0.699) was lower than that of the CRF-only model, and the specificity (0.524) was notably reduced, indicating a higher rate of false positives. These findings underscore the potential of the ECG-only model; however, its standalone utility in T2DM risk prediction remains limited. This emphasizes the need for integration with complementary data modalities.

Building on this baseline, the integration of ECG features with CRFs in the ECG-DiaNet model significantly enhanced the T2DM risk prediction performance compared to the CRF-only model. ECG-DiaNet achieved the highest AUROC of 0.845 (95% CI: 0.8427–0.8451), representing a statistically significant improvement over the CRF-only model, as indicated by the DeLong test ($p$-value < 0.001). This finding underscores the added discriminative value of ECG-derived features in the model, capturing complementary signals associated with T2DM that may not be reflected in CRFs alone. Beyond the AUROC, ECG-DiaNet demonstrated enhanced sensitivity (0.728) and specificity (0.782) compared with the CRF-only model, suggesting its ability to detect both true-positive cases and reduce false positives. Notably, the AUPRC (0.732) showed a marked improvement over the CRF-only model, emphasizing the utility of ECG features in refining predictions for high-risk individuals. The addition of ECG features also contributed to better model calibration, as evidenced by the lowest Brier score (0.173) among the evaluated models. A lower Brier Score indicates improved accuracy in probability predictions, meaning the predicted risks more closely align with the actual outcomes. This enhancement is crucial for clinical decision-making, as well-calibrated models provide more reliable risk assessments. Overall, the statistically significant improvement in performance metrics highlights the synergistic value of integrating ECG features with CRFs in multimodal approaches for T2DM risk prediction. The ROC, calibration, and AUPRC curves for the ECG-DiaNet model are depicted in Figures 2. These findings establish the potential of ECG-DiaNet as an effective non-invasive tool, providing a meaningful improvement in T2DM risk prediction.

**Table 2.** Performance of models in the 5-fold cross-validation for T2DM detection. The performance metrics are presented as mean values with standard deviations (SD) across 1,000 bootstraps for each test fold. The best results are highlighted in bold, and the second-best results are underlined.

| Model | AUROC (95% CI) | Sensitivity | Specificity | PPV | NPV | F1-Score | AUPRC |
|---|---|---|---|---|---|---|---|
| CRF-Only | 0.9841 (0.9839–0.9843) | 0.9484 ± 0.0166 | 0.9612 ± 0.0191 | 0.9668 ± 0.0162 | 0.9401 ± 0.0191 | 0.9574 ± 0.0125 | 0.9873 ± 0.0059 |
| ECG-Only | 0.8433 (0.8425–0.8440) | 0.8059 ± 0.0311 | 0.6951 ± 0.0597 | 0.7596 ± 0.0397 | 0.7510 ± 0.0362 | 0.7812 ± 0.0253 | 0.8665 ± 0.0320 |
| ECG-DiaNet ( ECG + CRFs) | 0.9852 (0.9851–0.9854) | 0.9514 ± 0.0170 | 0.9645 ± 0.0159 | 0.9695 ± 0.0136 | 0.9436 ± 0.0195 | 0.9602 ± 0.0119 | 0.9857 ± 0.0114 |

**Note:** CRF-Only: Age, sex, BMI, waist size, systolic and diastolic blood pressure. ECG-Only: ECG-derived features only. ECG-DiaNet (ECG + CRFs): Integration of ECG-derived features and CRFs.
**AUROC [95% CI]:** Area Under the Receiver Operating Characteristic Curve [95% Confidence Interval]. **SD:** Standard Deviation. **PPV:** Positive Predictive Value. **NPV:** Negative Predictive Value. **AUPRC:** Area Under the Precision-Recall Curve.

**Risk Stratification and Reclassification Analysis**

To further evaluate the impact of integrating ECG features with CRFs on T2DM risk prediction, we conducted a risk stratification analysis of the longitudinal cohort. Participants were divided into low-, medium-, and high-risk groups based on the quartiles of their predicted risk scores from both the CRF-only and ECG-DiaNet models. This stratification allowed for a comparison of the effectiveness of the models in identifying individuals at varying levels of risk, with a particular focus on the high-risk group, where early intervention is most critical. As illustrated in Figure 3, the PPV, which measures the proportion of positive predictions that are true positives, improved for the high-risk group when the ECG features were incorporated into the predictive model. Specifically, the ECG-DiaNet model achieved a PPV of 78% for the high-risk group compared to 72% for the CRF-only model. This enhancement indicates that the addition of ECG features increases the precision of the model in correctly identifying individuals who will develop T2DM, thereby improving the targeted prevention strategies. In contrast, the medium-risk group exhibited a slight reduction in PPV, while the low-risk group maintained very low PPV values in both models, suggesting a minimal risk of T2DM development. The increase in the PPV and reclassification metrics in the high-risk group has important clinical implications. By more accurately identifying individuals at the highest risk of developing T2DM, the ECG-DiaNet model facilitates targeted interventions such as intensive lifestyle modifications or preventive pharmacotherapy.

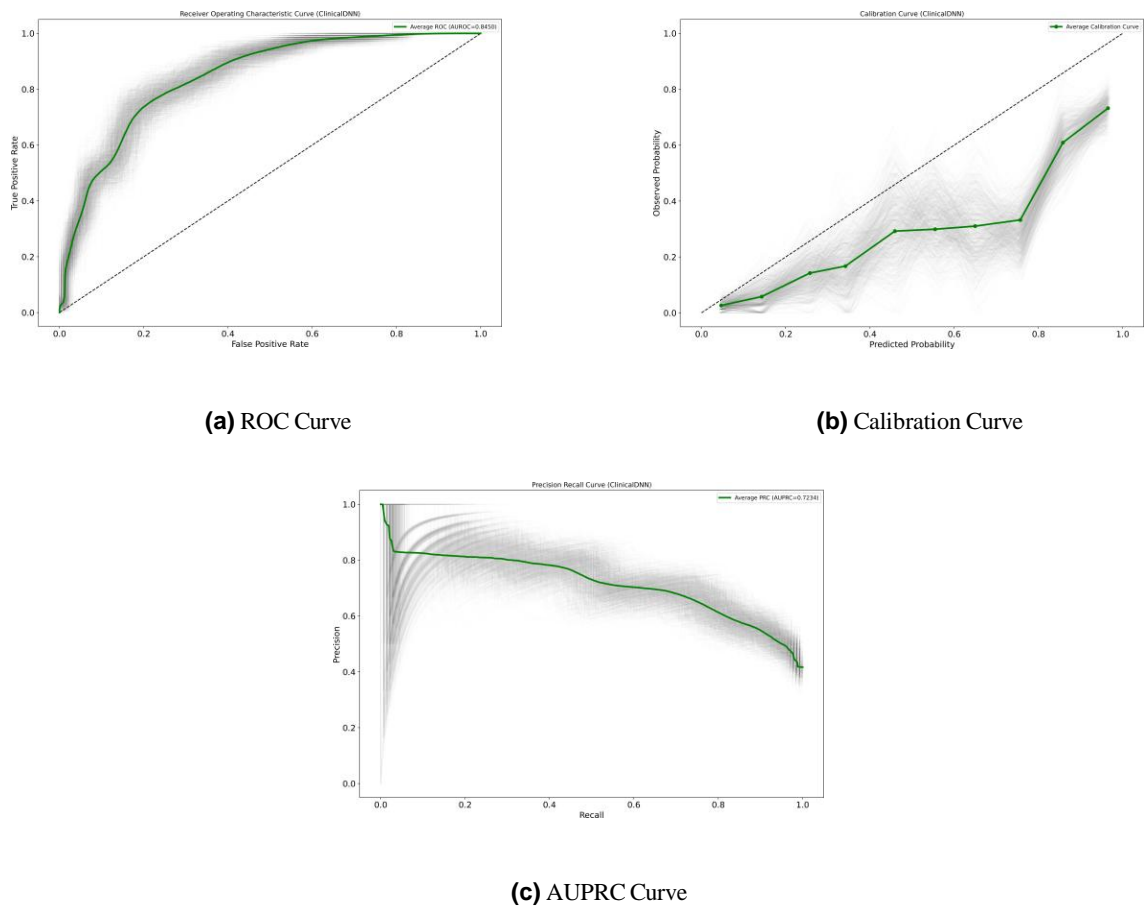

**(a)** ROC Curve

**(b)** Calibration Curve

**(c)** AUPRC Curve

**Figure 2.** ROC, precision-recall, and calibration curves of the ECG-DiaNet Multimodal Model Performance-based model. The green lines represent the average performance, while the grey shading illustrates the variability across 1000 bootstrap samples.

The enhanced stratification performance of the ECG-DiaNet model was further supported by significant improvements in reclassification metrics. The Net Reclassification Improvement (NRI) was $0.0153$ ($p < 0.001$), while the category-free NRI (cNRI) reached $0.5524$ ($p < 0.001$). Notably, the cNRI value of $0.5524$ indicates that incorporating ECG features led to a net improvement of over 55% in correctly reclassifying individuals into appropriate risk categories. Although the NRI mean appears modest, its statistical significance suggests that this improvement is unlikely to be due to random variation. The Integrated Discrimination Improvement (IDI) of $0.0482$ ($p = 0.0099$) further underscores the model's enhanced ability to discriminate between individuals who did and did not develop T2DM, reflecting improved overall sensitivity without

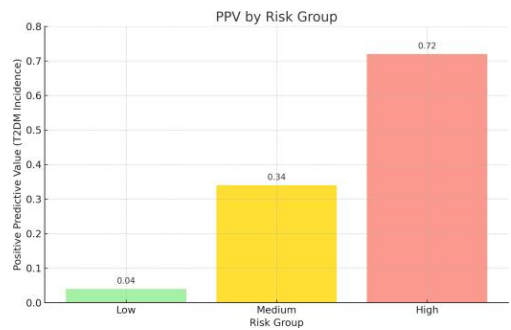

**(a)** Positive Predictive Value (PPV) by risk category (Low, Medium, High) for the CRF-only model.

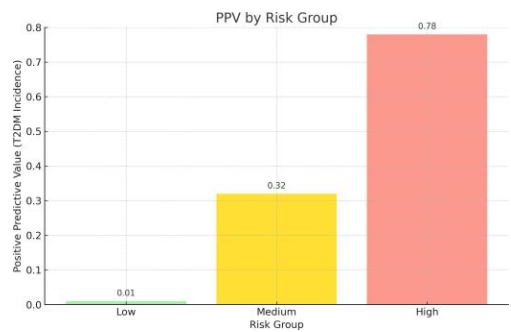

**(b)** Positive Predictive Value (PPV) by risk category (Low, Medium, High) for the ECG-DiaNet model.

**Figure 3.** Comparison of Positive Predictive Value (PPV) Across Risk Categories for the CRF-Only and ECG-DiaNet Models. The left panel (a) depicts PPV stratified by risk group for the CRF-only model, while the right panel (b) shows PPV for the ECG-DiaNet model. Incorporating ECG-derived features significantly enhances the PPV, particularly in the high-risk category.

compromising specificity. Collectively, these findings highlight the value of integrating ECG-derived features into T2DM risk prediction models and demonstrate the potential of ECG-DiaNet as a non-invasive tool for improving patient risk stratification and care.

## Discussion

In this study, we developed and evaluated ECG-DiaNet, a multimodal neural network that integrates ECG features with CRFs, to enhance the detection and risk prediction of T2DM. Our findings demonstrate that combining ECG data with traditional risk factors significantly improves the predictive performance over models using either modality alone.

### Main Findings

The primary finding of this study was that the integration of ECG features with CRFs in the ECG-DiaNet model significantly enhanced the predictive accuracy for T2DM risk prediction. In the longitudinal risk prediction evaluation, ECG-DiaNet achieved an AUROC of 0.845, which was significantly higher than that of the model that used known risk factors, including age, sex, BMI, waist circumference, and systolic and diastolic blood pressure, which obtained an AUROC of 0.8079. The integration of ECG features improved the sensitivity and specificity, with ECG-DiaNet attaining values of 69.20% and 81.83%, respectively, compared to 67.77% and 81.04% for the CRF-only model. Reclassification metrics further substantiated the added value of ECG features. The cNRI was 0.5524, indicating a significant net improvement in risk categorization. These results suggest that ECG-derived features capture complementary physiological information related to T2DM risk that is not fully represented by traditional clinical risk factors. The modest performance of the ECG-only model underscores that, while ECG features alone are informative, their predictive power is substantially amplified when combined with CRFs.

Beyond discrimination, calibration is an important aspect of the performance of a predictive model. Calibration refers to the agreement between the predicted probabilities and observed outcomes. The Brier score, a widely used measure for assessing calibration, indicates better calibration and overall accuracy of probability predictions when lower scores are obtained[12]. In this study, the ECG-DiaNet model achieved the lowest Brier score of 0.1716, providing more accurate estimates of actual risk. Good calibration is essential for clinical applicability, as it ensures that the predicted probabilities are reliable for decision-making. A well-calibrated model such as ECG-DiaNet allows healthcare providers to assess an individual's risk with greater precision, supporting tailored interventions and efficient resource allocation.

Our findings align with and expand upon those of previous research demonstrating the utility of ECG features in predicting cardiometabolic diseases. While earlier studies have demonstrated the feasibility of using ECG features to detect prevalent T2DM cases, our work uniquely contributes by focusing on predicting incident T2DM risk through a multimodal approach in a Middle Eastern population. Previous models, such as those developed by Lin et al. and Swapna et al.[9,10] have shown that machine learning techniques can effectively utilize ECG data to identify existing diabetes cases, highlighting the potential of ECGs as a non-invasive diagnostic tool. Kim et al. advanced this field by investigating the predictive value of ECG features in T2DM long-term risk evaluation in a South Korean cohort[13]. Their findings demonstrated the potential of ECG-derived patterns in forecasting future diabetes risk over a median follow-up period of 11 years. However, the generalisability of their findings to other populations remains uncertain because of differences in genetic, environmental, and lifestyle factors. In contrast, our study is among the first to evaluate the long-term predictive value of ECG-derived features for incident T2DM in the Middle Eastern population. Moreover, unlike earlier approaches that primarily utilized unimodal data, our multimodal approach integrates ECG features with CRFs, resulting in more comprehensive risk prediction. By integrating ECG features with CRFs, including age, sex, BMI, waist circumference, and blood pressure, we developed the ECG-DiaNet model, which leveraged the strengths of both data types. Achieving an AUROC of 0.845 in predicting incident T2DM, ECG-DiaNet shows promising potential for use in non-invasive and readily available data. Our findings demonstrate that the inclusion of ECG features provides complementary value to traditional risk factors. While CRFs capture established demographic and clinical predictors of T2DM, ECG data reflect the underlying cardiovascular alterations that may precede or coincide with metabolic changes. This multimodal approach addresses the multifactorial nature of T2DM risk more comprehensively than unimodal models. Furthermore, our study's focus on the Middle Eastern population addresses a gap in the literature, as most prior research has been conducted in Western or East Asian populations. Given the high prevalence of T2DM in the Middle East and the unique genetic and environmental factors at play, developing and validating predictive models in this context is crucial for effective disease management and prevention strategies.

### Clinical Implications

The enhanced predictive performance of the ECG-DiaNet has significant real-world implications. Early and accurate identification of individuals at high risk of T2DM is critical for implementing targeted preventive strategies that can delay or prevent disease onset. Moreover, the non-invasive nature of ECGs, coupled with their widespread availability, makes the implementation of ECG-DiaNet feasible in diverse clinical settings, including primary care and community health centers.

**Table 3.** T2DM risk prediction performance of various models, evaluated using 1000 bootstrapped samples from the longitudinal test set (Mean ± SD)

| Model | AUROC (95% CI) | Sensitivity | Specificity | PPV | NPV | F1-Score | AUPRC | Brier Score |
|---|---|---|---|---|---|---|---|---|
| CRF-Only | 0.8217 (0.8205–0.8229) | 0.764 ± 0.036 | 0.744 ± 0.028 | 0.620 ± 0.037 | 0.852 ± 0.024 | 0.684 ± 0.030 | 0.699 ± 0.038 | 0.199 ± 0.017 |
| ECG-Only | 0.6747 (0.6730–0.6764) | 0.699 ± 0.038 | 0.524 ± 0.031 | 0.445 ± 0.032 | 0.760 ± 0.032 | 0.543 ± 0.031 | 0.528 ± 0.043 | 0.253 ± 0.011 |
| ECG-DiaNet (ECG + CRFs) | 0.845 (0.8427–0.8451) | 0.728 ± 0.037 | 0.782 ± 0.026 | 0.647 ± 0.037 | 0.840 ± 0.024 | 0.684 ± 0.030 | 0.732 ± 0.039 | 0.173 ± 0.014 |

**Note:** CRF-Only: Age, sex, BMI, waist size, systolic and diastolic blood pressure. ECG-Only: ECG-derived features only. ECG-DiaNet (ECG + CRFs): Integration of ECG-derived features and CRFs. **AUROC [95% CI]:** Area Under the Receiver Operating Characteristic Curve [95% Confidence Interval]. **SD:** Standard Deviation. **PPV:** Positive Predictive Value. **NPV:** Negative Predictive Value. **AUPRC:** Area Under the Precision-Recall Curve.

Integrating this model into electronic health record systems could facilitate automated risk assessments during routine visits, enabling timely interventions without adding a significant burden to clinicians or patients. From a public health perspective, adopting the ECG-DiaNet could help reduce the incidence and overall burden of T2DM. Early intervention not only improves individual health outcomes but also has the potential to lower healthcare costs associated with managing advanced disease and its complications. Furthermore, the model supports personalized medicine by tailoring risk assessments based on individual physiological profiles.

**Limitations and Future Directions**

Despite these promising findings, several limitations should be acknowledged. This study was conducted using data from a single geographical region, which may limit the generalisability of the results to other populations, particularly across different age groups and ethnicities. Additionally, the longitudinal cohort used for risk prediction testing consisted of a relatively small sample size, potentially affecting the robustness and generalizability of the findings. The retrospective nature of the study may have introduced biases related to data collection and patient selection, underscoring the need for prospective validation in real-world clinical settings. The current model also relies solely on ECG data and a limited set of clinical risk factors, without incorporating other potentially valuable non-invasive biomarkers, such as retinal imaging, genomics, or wearable sensor data, which could enhance its predictive power and comprehensiveness. Furthermore, while ECG-DiaNet demonstrates good predictive performance, the interpretability of deep learning models remains a challenge, necessitating further investigation into explainable AI (XAI) techniques to ensure clinical utility and trust. Finally, ethical and data-sharing considerations present additional challenges, as sensitive health data may not be publicly available despite the importance of reproducibility and external validation. Future studies should address these limitations by validating the model in larger, more diverse cohorts, including varied age groups and ethnicities, integrating additional data sources, improving interpretability, and developing secure data-sharing frameworks to ensure broader applicability and compliance with ethical guidelines.

## Conclusion

This study demonstrated that integrating ECG-derived features with known risk factors significantly enhances the prediction of future T2DM risk. The ECG-DiaNet model outperformed unimodal models by providing improved risk stratification. The model provides a more comprehensive assessment of T2DM risk by capturing complementary physiological information related to cardiac function. The real-world implications of our findings are noteworthy. ECG-DiaNet is a practical, non-invasive tool that can be seamlessly integrated into clinical workflows, enabling early identification and personalized intervention strategies. By enhancing the risk prediction, the model has the potential to improve patient outcomes and alleviate the healthcare burden associated with T2DM. This study underscores the importance of multimodal approaches in disease risk prediction, paving the way for future research aimed at refining and validating models across diverse populations and settings. As precision medicine continues to evolve, integrating diverse data sources, such as ECGs, with traditional risk factors offers a promising avenue for advancing preventive healthcare and addressing the global challenge of T2DM.

## Methods

**Ethical Approval**

This research was carried out in compliance with Qatar's Ministry of Public Health regulations. This work was approved by the Institutional Review Board of QBB in Qatar and used a de-identified dataset from the QBB. The QBB dataset is not publicly available, in compliance with the Qatar Biobank data-sharing policy.

**Data collection and preprocessing**

The data used in this study was collected from the QBB. The details of the data collection protocol adopted by the QBB was described in[2,11]. In brief, participants were invited to Staff nurses interviewed the QBB to collect their background history. Then Multiple laboratory tests and imaging, such as ECG scans and retinal images, were collected. The data for this study included the clinical and ECG features extracted by the QBB team. In addition to ECG data, clinical known risk factors are selected as reported by the litrature[14–21], six key factors were selected for analysis: non-invasive factors (age, gender, BMI, waist size, systolic and diastolic BP).

The ECG dataset comprised features derived from three non-consecutive 30-second resting ECG recordings for each participant, extracted by the QBB team. Further details regarding the data collection process were provided by Kuwari et al.[2,11]. For this study, features were selected based on their clinical relevance to cardiac electrophysiology and potential association with T2DM and related cardiovascular abnormalities. These features included the QRS duration, QT interval, corrected QT interval, PR interval, average RR interval, and T-wave axis. A summary of these features and their clinical implications is presented in Table 4. These features were chosen to capture diverse aspects of cardiac electrophysiology, offering insights into the cardiovascular patterns associated with T2DM risk.

The ECG features and CRFs were processed using a structured preprocessing pipeline to handle missing values, outliers, and inconsistencies. Missing continuous variables were imputed with their mean values, whereas categorical variables were imputed with their mode. Outliers detected using interquartile range analysis were addressed by replacing the extreme values with the mean of the respective features. To standardize the data and support effective model training, all features were scaled using MinMaxScaler.

**Table 4.** ECG Features and Their Clinical Significance

| Feature | Description | Clinical Significance |
| --- | --- | --- |
| **QRS Duration** | Duration of the QRS complex, reflecting the time required for ventricular depolarization. | Prolonged QRS duration is linked to an increased risk of cardiac mortality and heart failure[22]. |
| **QT Interval** | Total time of ventricular depolarization and repolarization. | Abnormal QT intervals are associated with a higher risk of ventricular arrhythmias and sudden cardiac death[23]. |
| **Corrected QT Interval (QTc)** | QT interval adjusted for heart rate variability using Bazett's formula. | Provides a more accurate assessment of ventricular repolarization abnormalities, which are linked to arrhythmia risk[?]. |
| **PR Interval** | Time from the onset of atrial depolarization to the beginning of ventricular depolarization. | Abnormal PR intervals may indicate atrioventricular conduction disturbances, such as first-degree AV block[24]. |
| **Average RR Interval** | Average interval between successive R-waves, reflecting heart rate and variability. | Serves as a marker of autonomic nervous system function and cardiovascular health[25]. |
| **T-Wave Axis** | Electrical axis of the T-wave, indicating the direction of ventricular repolarization. | Deviations in T-wave axis can signal ischemic heart disease or ventricular hypertrophy[26]. |

## Study design

The dataset considered in this study comprises two distinct cohorts derived from the QBB, enabling robust model development and validation. The cohort included baseline cross-sectional data and a longitudinal follow-up cohort to evaluate T2DM risk prediction. The development cohort consisted of 4,500 participants with cross-sectional baseline data. After applying exclusion criteria (e.g. missing ECG data or participants with pre-existing prediabetes), the final development cohort comprised 2043 participants, of whom 1107 had confirmed T2DM diagnoses and 937 were non-diabetic controls (see Figure 1). The classification of T2DM cases and controls was conducted with the assistance of the QBB medical practitioners and nurses. The Longitudinal test cohort included 1500 participants with baseline data and a median follow-up duration of five years. Participants with T2DM at baseline were excluded to focus on predicting incident T2DM ( Figure 1).The final longitudinal cohort included 395 participants, of whom 140 developed T2DM during the follow-up period. The diagnostic criteria included HbA1c ≥ 6.5%, fasting plasma glucose ≥ 126 mg/dL, or self-reported use of diabetes medications.

## Experiment Setup and Model Development

We designed three experiments using unimodal and multimodal data configurations to evaluate the contributions of ECG data, CRFs, and their integration in predicting T2DM. The first configuration, the ECG-only model, employed a deep neural network (DNN) to process ECG-derived features exclusively. The second configuration, the CRF-only model, analyzed CRFs using a similar DNN architecture. The third configuration, the ECG+CRF multimodal model (ECG-DiaNet ), combined ECG features and CRFs into a unified deep learning model, leveraging their complementary strengths to enhance prediction accuracy.

### *ECG-Only and CRF-Only Networks*

The DNN architecture for the ECG-only model consisted of three fully connected layers with ReLU activations applied to the first two layers to introduce nonlinearity. Dropout layers were incorporated to mitigate overfitting, and the final layer produced a single prediction for T2DM risk using a sigmoid activation function. The input features included heart rate variability, QRS complex metrics, QT interval, PR interval, average RR interval, and T-wave axis, all derived from 30-second resting ECG recordings. These features were chosen for their clinical relevance to cardiac electrophysiology and their potential association with T2DM and related cardiovascular abnormalities.

Similarly, the CRF-based model used the same DNN architecture to process clinical and demographic variables. The model was designed to capture non-linear interactions among these diverse risk factors, reflecting the multifactorial nature of T2DM. Dropout layers were used to ensure model generalizability, and the final output layer provided a binary prediction for T2DM risk. This configuration emphasized systemic metabolic and demographic characteristics, serving as a complementary counterpart to the ECG-based model.

### *ECG+CRF Multimodal Model (ECG-DiaNet )*

The ECG-DiaNet model combined the feature extraction capabilities of the ECG-only and CRF-only models into a multimodal predictive model. Feature vectors from the two models were concatenated and passed through additional fully connected layers designed to learn synergistic patterns across the two data modalities. This early fusion approach enabled the model to integrate complementary information from cardiac electrophysiology and clinical risk profiles, enhancing its ability to capture the complex interplay of cardiovascular and metabolic factors underlying T2DM risk. The final layer applied a sigmoid activation function to produce a binary classification for T2DM.

## Model Training and validation

The model development process adhered to a rigorous training and validation protocol to ensure robustness and generalizability. Five-fold cross-validation was employed using the development cohort, which consisted of cross-sectional data from participants with and without T2DM. The dataset was randomly partitioned into five subsets, with four subsets used for training and the fifth reserved for validation in each iteration. To ensure stability and confidence in performance estimates, this process was repeated 1,000 times using bootstrapping. Hyperparameter tuning was conducted during cross-validation to optimize the architecture, learning rate, and regularization parameters of the models. Once the models were optimized, they were trained on the entire development cohort. The trained models were then evaluated on a longitudinal test cohort comprising non-diabetic individuals at baseline, assessing their risk of developing T2DM over a follow-up period. To avoid data leakage, there was no overlap between the training and evaluation sets, ensuring a robust assessment of the models' ability to predict long-term T2DM risk. The binary cross-entropy loss of the model was optimized using the Adam optimizer[27] with a learning rate of 5e-4. The models were trained with weight decay set to 5e-6 for 20 epochs with a batch size of 16.

## Risk Prediction and Statistical Analysis

The models were evaluated on the longitudinal test set to estimate the risk of developing T2DM during follow-up visits based on baseline data. For each metric, 1,000 bootstrap samples were generated, and the mean and standard errors were reported. Model performance was primarily assessed using the area under the receiver operating characteristic curve (AUROC), along with 95% confidence intervals. Additional metrics included the area under the precision-recall curve (AUPRC), sensitivity, specificity, positive predictive value (PPV), and negative predictive value (NPV). Calibration was evaluated using the Brier score, which quantifies the accuracy of predicted probabilities, with lower scores indicating better calibration. Calibration curves were also plotted to assess the agreement between predicted and observed risks, ensuring the model's reliability for clinical decision-making.

## Risk Stratification Analysis

To evaluate the effectiveness of integrating ECG-derived scores into a traditional CRF-only model for predicting the onset of T2DM, we performed a risk stratification analysis. We classified participants into distinct risk categories based on their predicted scores. The predicted probabilities were calculated using baseline data, and the actual incidence of T2DM within each risk group was compared over a 5-year follow-up period. Specifically, we compared the risk stratification outcomes between the CRF-only model and the ECG-DiaNet model, which combines ECG-derived features with traditional CRFs.

Risk stratification was conducted by categorizing participants into three distinct risk groups: Low, Medium, and High. The categorization was based on quantile thresholds, with the lowest 20% of scores categorized as Low Risk, the middle 20-80% as Medium Risk, and the highest 20% as High Risk. The incidence rates, expressed as positive predictive values (PPV), were calculated for each risk group. This approach enabled us to evaluate how effectively each model distinguished participants based on their risk of developing T2DM. For visualization, PPV was calculated for each risk group under both models. The results were displayed as bar charts, where the distribution and PPV of each risk group could be directly compared.

## Funding Statement

Open Access funding provided by Qatar National Library.

## Data Availability

The datasets generated and/or analyzed during the current study are not publicly available due to a non-disclosure agreement. However, they can be accessed through an application to Qatar Biobank. Requests can be submitted online and are subject to approval by the institutional review boards of Qatar Biobank. For further assistance, users may contact the corresponding author or send an email to qbbresearch@qf.org.qa to initiate the data access request.

## Authors' contributions

F.M. contributed to the conceptualization of the study. F.M. and Z.S. involved in data collection and labeling. F.M. performed all simulations and contributed to the writing of the paper. Z.S. contributing to the review of the manuscript. F.M. and Z.S. supervised the study. All authors approved the manuscript for publication and agreed to be accountable for all aspects of the work.

## Corresponding authors

Correspondence to Farida Mohsen (famo39885@hbku.edu.qa) and Zubair Shah (zshah@hbku.edu.qa).

## Ethics declarations

### Ethics approval and consent to participate
This study was approved by the Institutional Review Board of the Qatar Biobank and conducted in accordance with Qatar's Ministry of Public Health regulations. Informed consent was obtained from all participants, and only de-identified data were used.

### Consent for Publication
All authors have read and approved the final manuscript for submission.

### Conflict of Interest declaration
The authors declare that they have no conflict of interest.

## Declaration of generative AI and AI-assisted technologies in the writing process

During the preparation of this work, the author(s) used ChatGPT-4 in order to proofread and improve the language and readability. After using this tool/service, the author(s) reviewed and edited the content as needed and take(s) full responsibility for the content of the publication.